\documentclass[runningheads]{llncs}
\usepackage[T1]{fontenc}
\usepackage[subtle]{savetrees}
\usepackage{caption} 
\usepackage{graphicx,verbatim}
\usepackage{amsmath}
\usepackage{amssymb}
\usepackage{multirow}
\usepackage{placeins}   
\usepackage{float}
\usepackage[hidelinks]{hyperref}
\usepackage{subcaption} 
\usepackage{xcolor}
\usepackage{placeins}

\usepackage{subcaption}
\usepackage{xcolor}

\begin{document}
\title{Spectral Priors vs. Attention: \\Investigating the Utility of Attention\\ Mechanisms in EEG-Based Diagnosis}
\titlerunning{Spectral Priors vs Self Attention}
%

\author{Tawsik Jawad \and
Gowtham Atluri \and
Vikram Ravindra}

\institute{
University of Cincinnati, Cincinnati, OH, USA\\
\email{jawadkk@mail.uc.edu, \{atlurigm, ravindvm\}@ucmail.uc.edu}
}
  
\maketitle              

\begin{abstract}
Electroencephalograph (EEG) time-series signals are characterized by significant noise and coarse spatial resolution, which complicates the classification of neuro-degenerative diseases. Even the state-of-the-art deep learning architectures struggle to distinguish between healthy controls and diseased subjects—or between different disease types due to high inter-group similarity. 

In this paper, we show that a spectrally selective approach to feature construction  enhances class separability. By isolating signal strengths within the primary brainwave bands— $\delta, \theta, \alpha, \beta, \gamma$ — we transform high-dimensional raw data into high-value spectral features.

Our results demonstrate that in small clinical datasets: a) features derived from frequency and time-frequency domain allow traditional machine learning models to match or exceed the performance of state-of-the-art deep learning models, b) Attention mechanism is unable to distill the stable feature signatures that characterize healthy neural activity in both resting and task EEGs, and c) the limitations of attention-based models in finding relevant spectral features appear to be robust in that providing frequency-selective time-domain input do not appreciably improve their performance.
We validate our methodology across three open-source resting EEG datasets, and one task EEG dataset providing robust empirical evidence for our claims. 


\end{abstract}

\section{Introduction}

Electroencephalography (EEG) is a non-invasive neuroimaging data modality that measures the electrical activity in the brain. EEG recordings are comprised of an array of electrodes placed on the surface of the skull. EEGs measure brain activity in terms of minute voltage changes between pairs of electrodes. EEG signals are widely used to understand, diagnose, evaluate, and monitor the brain activity and cognitive ability in neuro-degenerative diseases. They are also used extensively as input modalities by Brain-Computer Interfaces (BCIs).

Time-series classification of EEG data is of particular clinical interest for diagnosing and predicting the onset of different neurological diseases and disorders (such as Alzheimer's Disease, Parkinson's Disease, Dementia, and Attention Deficit Disorders). 
Attention-based Transformer architectures have been proposed with the goal of levaraging temporal dependence in the signals \cite{c04,c05,c06}. However, EEG signals are known to exhibit significant inter-subject variability and non-stationary temporal dynamics. While Transformers excel at modeling long-range dependencies in quasi-stationary sequences (like text data), they lack the specific inductive bias required to disentangle the overlapping oscillatory components inherent in neuro-physiological data. Recent approach by Yu et al \cite{c27} have engineered a centralized token distribution strategy to align with the globally distributed discriminative signals in medical time-series.

In this paper, we show that traditional classification approaches, such as Quadratic Discriminant Analysis and Random Forests operating on spectral features are comparable, and sometimes outperform state-of-the-art attention-based architectures in diagnosing different neuropathologies in small EEG datasets. We also demonstrate that providing these spectral features as input do not result in a significant improvement in performance of attention-based architectures, suggesting that the attention mechanism suffers from a lack of temporal salience in quasi-stationary signals, where discriminative biomarkers are distributed globally across the recording rather than localized in the discrete temporal events that self-attention is designed to prioritize. Methodically, a light-weight framework extracting neuro pathology informed biomarkers provide stronger decision boundaries in the benchmark EEG datasets.

\section{Methodology}

Given a dataset of EEG data drawn from a population and associated labels (ex: healthy v/s disease), our goal is to construct relevant features with high discriminatory power. We initiate the description of our approach by introducing relevant terminology and techniques.

\subsection{Feature Construction}
\label{subsec:prelim}
 Let an EEG session be represented by a matrix $X \in \mathbb{R}^{C\times T}$ where $C$ represents the number of channels and $T$ represents the number of time-points. After downsampling all the datasets to 256 Hz, we feed signal strengths at five frequency bands (PSD values) and spectral frequency measures calculated via Welch FFT to Machine Learning models. We band pass the original signal to keep the time-signal representing frequency bands from 0.5-45 Hz. We denote the sampling rate as $f_s$. From each such session, we a) center each time series, b) extract spectral features using the Welch method of Fast-Fourier Transform (FFT) applied to get signal strengths at frequency bands and spectral frequency values in frequency domain, and c) apply Discrete Wavelet Transform (DWT) to capture the non-stationary frequency components of time-signal and get the relative band energies in the time-frequency domain.


\textbf{Welch-based FFT band powers:}
For each channel 
$\tilde{x}_c$ 
is our centered zero-mean time series. 
For frequency resolution of time signal, we estimate the Power Spectral Density (PSD) via Welch’s method using a Hann window with overlapping segments \cite{c13} and denote it by $\widehat{S}_c(f)$. 
For each band $b=[f_b^{\min}, f_b^{\max}]$, the absolute band power is $P_{c,b}^{\mathrm{abs}}=\int_{f_b^{\min}}^{f_b^{\max}} \widehat{S}_c(f)\,df$. The relative band powers over different frequency bands of interest are given as $P_{c,b}^{\mathrm{rel}}=\frac{P_{c,b}^{\mathrm{abs}}}{\sum_{b'\in\mathcal{B}} P_{c,b'}^{\mathrm{abs}} + \varepsilon}$ (with $\varepsilon=10^{-12}$).
We compute the relative band power for the 5 canonical EEG bands
($\delta$: $0.5$--$4\,\mathrm{Hz}$, $\theta$: $4$--$8\,\mathrm{Hz}$, $\alpha$: $8$--$13\,\mathrm{Hz}$, $\beta$: $13$--$30\,\mathrm{Hz}$, $\gamma$: $30$--$45\,\mathrm{Hz}$)
and aggregate across channels to form the session-wise FFT feature $\mathrm{FFT}_b=\frac{1}{C}\sum_{c=1}^{C} P_{c,b}^{\mathrm{rel}}$.

Using the same Welch PSD $\widehat{S}_c(f)$ (restricted to $0.5$--$45$ Hz), we compute {per-channel mean frequency, median frequency, and spectral entropy}, and then average across channels. Denoting the normalized spectrum by $p_c(f)=\widehat{S}_c(f)/\int \widehat{S}_c(u)\,du$, spectral entropy is $H^{(c)}=-\int p_c(f)\ln p_c(f)\,df$, with the remaining summary statistics computed analogously from $p_c(f)$ and then channel-averaged.

\textbf{Discrete Wavelet Transform (DWT) band energies:}
For Non-Stationary Time Signals, the temporal component of frequencies which are not found from Welch-FFT are captured by DWT. For each channel $c\in\{1,\ldots,C\}$, we have $\tilde{x}_c\in\mathbb{R}^{T}$ as the per-channel DC-centered time series and $f_s$ is our sampling rate. We compute an $L$-level DWT (wavelet $\psi$, default \texttt{db4}), yielding approximation coefficients $cA_{c,L}$ (longer time resolution for low-frequency up to $f_s/2^{L+1}$) and detail coefficients $\{cD_{c,j}\}_{j=1}^{L}$ (shorter time-resolution for higher frequency components in dyadic(power of 2) frequency ranges). We define level energies by squared $\ell_2$ norms:
$
E^{A}_{c,L}=\lVert cA_{c,L}\rVert_2^2, \ 
 E^{D}_{c,j}=\lVert cD_{c,j}\rVert_2^2,\ \ j=1,\ldots,L.
$

Each detail level $j$ is associated with the nominal dyadic interval
$\mathcal{I}_j=(f_s/2^{j+1},\,f_s/2^{j})$.
To obtain energies in the canonical EEG bands $\mathcal{B}=\{\delta,\theta,\alpha,\beta,\gamma\}$, we map the DWT subband energies---those derived from the approximation coefficients $cA_{c,L}$ with support $I^{A}=\left(0,\frac{f_s}{2^{L+1}}\right)$, and from the detail coefficients $\{cD_{c,j}\}_{j=1}^{L}$ with supports $\mathcal{I}_j$---to each EEG band. We compute the fractional overlap between subband intervals and the target band limits for this mapping. This yields a deterministic mapping from DWT subbands to canonical bands (rather than a within-band PSD estimate). We then normalize within channel and average across channels to obtain session-wise features $\mathrm{DWT}_b$. We select $L$ per epoch as the largest admissible decomposition depth given the epoch length and wavelet support (PyWavelets \texttt{dwt\_max\_level}), capped at a maximum (here $L\le 8$). For example, with $f_s=256$ Hz and $T=256$ (1 s), \texttt{db4} yields $L=5$ so that, $cA_{c,L}$ = $f_s/2^{L+1}=4$ Hz, aligning the approximation cutoff with the upper edge of the $\delta$ band.

\emph{In all, our feature-set is comprised of relative PSD in each 5 frequency bands, relative wavelet strengths in those frequency bands, and spectral summary features (spectral mean frequency, spectral median frequency, and spectral entropy)}


\subsection{Classical Baselines and Spectral Feature Classification}
We train simple classifiers that operate on spectral features (described above) to output predicted labels. In our experiments, we choose Quadratic Discriminant Analysis (QDA) and Random Forest. 
QDA minimizes the total probability of misclassification by assigning each observation to the class with the highest posterior probability. It achieves this by calculating a quadratic decision boundary that minimizes the expected loss based on class-specific Gaussian distributions. Random Forest is an ensemble learning technique that works by building a many independent decision trees, each trained on a random subset of data and a random selection of features. It produces the final prediction using a majority vote. 
In the context of diagnosis with EEG data, we show that these classifiers match, and sometimes exceed many state-of-the-art attention-based deep learning models.

\section{Results}

\subsubsection{Datasets}

We demonstrate our results on a total of four publicly available datasets for different pathologies, including three resting state EEG datasets and one task EEG dataset. 
The APAVA \cite{c10}  dataset has 16 channels and 23 subjects, 12 with Alzheimers and 11 Healthy as labels. 
The TDBrain \cite{c11} dataset has 33 channels and we sample 50 subjects total where 25 have Parkinsons and 25 are Healthy Controls. 
ADFTD \cite{c12} has 88 subjects, 19 channels with three classes, where 23 are Dementia,  36 Alzheimers, and 29 Healthy controls.
For task-EEG, we integrated ADHD \cite{c27} dataset having 19 channels with 61 ADHD patients and 60 Healthy Controls

\subsection{Performance parity is observed between manually designed features on traditional classifiers and Attention-Based methods}

For each EEG session in each dataset, we compute spectral features as described in section \ref{subsec:prelim}. We partition the data into train, validation and test sets using the protocol established in Medformer~\cite{c04} to ensure a fair comparison. 
Features undergo Z-score normalization, where the mean and standard deviation are computed strictly on the training set and subsequently applied to the validation and test sets to prevent data leakage.

We evaluate two traditional classifiers—Quadratic Discriminant Analysis (QDA) and Random Forests (RF)—against three state-of-the-art attention-based architectures: Medformer, Reformer \cite{c08}, and EEG Conformer \cite{c22}. For the traditional classifiers, we utilize Principal Component Analysis (PCA) to reduce dimensionality, selecting hyperparameters via Grid Search on the validation set to maximize macro-F1. In QDA, we tune the retained PCA variance and the regularization parameter to account for multi-collinearity. For the Random Forest, we adjusted the number of estimators, maximum depth, and minimum samples required per leaf.

We compare end-to-end pipelines rather than isolated classifiers. Each pipeline — classical (spectral feature extraction followed by QDA or RF) and attention-based (raw time-series processed by Medformer, Reformer, or Conformer) — receives identical training data, identical train/validation/test splits, and is evaluated on identical test subjects with identical metrics. Each pipeline is permitted to use the input representation appropriate to its architecture: classical methods consume engineered spectral features, while attention-based methods consume the raw time-series their architectures are designed for. This reflects how these pipelines would be deployed in practice and is the standard protocol for comparing methods that differ in where feature construction occurs. Restricting classical methods to raw input, or restricting attention-based methods to engineered features, would not be a more rigorous comparison; it would be an artificial constraint that hobbles each method's best version of itself.

\begin{table}[!htbp]
\centering
\setlength{\tabcolsep}{7pt}
\renewcommand{\arraystretch}{1.12}
\caption{\textbf{Macro-averaged classification performance (\%) of traditional classifiers (QDA and Random Forest) and attention-based methods (Medformer, Reformer, and Conformer) on four EEG datasets.} For each classifier, macro-averaged metrics are reported per dataset. \emph{The highest standard deviation found for the ML models were 2.35 (RF in ADFTD) and 1.32 for (Conformer in ADHD).} We mark the best metric scores for each dataset in bold. Our results show that across the 4 EEG datasets, QDA has outperformed all other models in 3 datasets (RF outperforms QDA in TDBrain) with FFT DWT components aggregated on timestamps and channels. In general classical ML pipelines have comparable numbers with that of Transformer models in all the datasets. Among the Transformer models, TeCh produces strong numbers in the APAVA and TDBrain,  Medformer outperforms the other Transformer models in ADFTD and ADHD. The high performance metrics obtained for the TDBrain and APAVA datasets by RF and QDA respectively can be attributed to the choice of the evaluation test set. However, to ensure comparability across the methods, we have chosen to preserve the same train/test split as reported in the Medformer paper \cite{c04}.}
\label{tab:classic_baselines}
\begin{tabular}{llcccccc}
\hline
\textbf{Classifier} & \textbf{Dataset} &
\textbf{Acc} & \textbf{Prec} & \textbf{Rec} & \textbf{F1} & \textbf{AUROC} & \textbf{AUPRC} \\
\hline
\multirow{4}{*}{\textbf{QDA}}
& ADFTD & \textbf{68.42} & \textbf{63.89} & \textbf{65.00} & \textbf{61.19} & 67.60 & 57.71 \\
& APAVA & \textbf{100.0} & \textbf{100.0} & \textbf{100.0} & \textbf{100.0} & 100.0 & 100.0 \\
& TDBrain & 87.50 & 90.00 & 87.50 & 87.30 & 92.50 & 94.33 \\
 & ADHD   & \textbf{78.95} & \textbf{78.89} & \textbf{78.89} & \textbf{78.89} & \textbf{86.67} & \textbf{79.96} \\
\hline
\multirow{4}{*}{\textbf{RF}}
 & ADFTD   & 53.68 & 43.47 & 50.28 & 46.52 & 70.33 & \textbf{62.99} \\
 & APAVA  & 75.0 & 83.33 & 75.0 & 73.33 & 100.0 & 100.0 \\
 & TDBrain& \textbf{100.0} & \textbf{100.0} & \textbf{100.0} & \textbf{100.0} & \textbf{100.0} & \textbf{100.0} \\
 & ADHD   & 72.63 & 75.44 & 73.44 & 72.23 & 82.0 & 78.27 \\
\hline\hline
\multirow{4}{*}{\textbf{TeCh}}
& ADFTD   & 54.54 & 53.02 & 49.25 & 48.84 & 68.67 & 50.62 \\
& APAVA  & 86.86 & 86.85 & 86.10 & 86.30 & 94.02 & 93.79 \\
& TDBrain & 93.21 & 93.39 & 93.21 & 93.20 & 98.68 & 98.72 \\
& ADHD & 71.10 & 71.78 & 72.11 & 71.0 & 82.43 & 81.45 \\
\hline

\multirow{4}{*}{\textbf{Medformer}}
& ADFTD   & 53.27 & 51.02 & 50.71 & 50.65 & \textbf{70.93} & 51.73 \\
& APAVA  & 78.74 & 81.11 & 75.40 & 76.31 & 85.59 & 84.39 \\
& TDBrain & 89.62 & 89.68 & 89.62 & 89.62 & 96.41 & 96.51 \\
& ADHD   & 73.35 & 73.32 & 73.33 & 73.10 & 80.60 & 71.41\\
\hline
\multirow{4}{*}{\textbf{Reformer}}
& ADFTD   & 50.78 & 49.64 & 49.89 & 47.94 & 69.17 & 51.73 \\
& APAVA  & 78.70 & 82.50 & 75.00 & 75.93 & 73.94 & 76.04 \\
& TDBrain & 87.92 & 88.64 & 87.92 & 87.85 & 96.30 & 96.40 \\
& ADHD   & 71.72 & 72.98 & 73.16 & 71.70 & 82.20 & 76.67\\
\hline
\multirow{4}{*}{\textbf{Conformer}}
& ADFTD   & 47.02 & 43.78 & 44.12 & 43.42 & 63.00 & 45.26 \\
& APAVA  & 85.10 & 86.56 & 82.88 & 83.65 & 87.99 & 87.38 \\
& TDBrain & 87.10 & 87.89 & 87.10 & 87.03 & 96.43 & 96.60 \\
& ADHD   & 71.59 & 73.54 & 73.25 & 71.52 & 83.62 & 75.42\\
\hline
\end{tabular}
\end{table}

The results, summarized in Table \ref{tab:classic_baselines} show that on multiple metrics, QDA achieves superior performance in ADFTD, APAVA, and ADHD. Random Forest provides the best results on the TDBrain dataset. 
We note that the core premise of the Transformer architecture is the existence of latent temporal ``events'' that require dynamic weighting. In the context of resting-state EEG (ADFTD, APAVA, and TDBrain datasets), however, the biomarkers of interest are primarily oscillatory power and phase-coupling \cite{c21,c26}. These measures are inherently stationary over the analyzed windows. By treating the EEG as a sequence of discrete tokens, Transformers ignore the underlying continuous physics of the signal, attempting to ``attend'' in a temporal domain that lacks salient landmarks. While similar considerations are relevant in task EEG (ADHD dataset), there are some event-related epochs that attention mechanisms can leverage. This explains the competitive precision, AUROC, and AUPRC by Transformer models. Finally, we note that QDA and RF perfectly classify APAVA and TDBrain, suggesting strong within-class grouping in these two datasets.

\begin{figure}[!htbp]
\centering
\begin{subfigure}[t]{0.48\linewidth}
    \centering
    \includegraphics[scale=0.60]{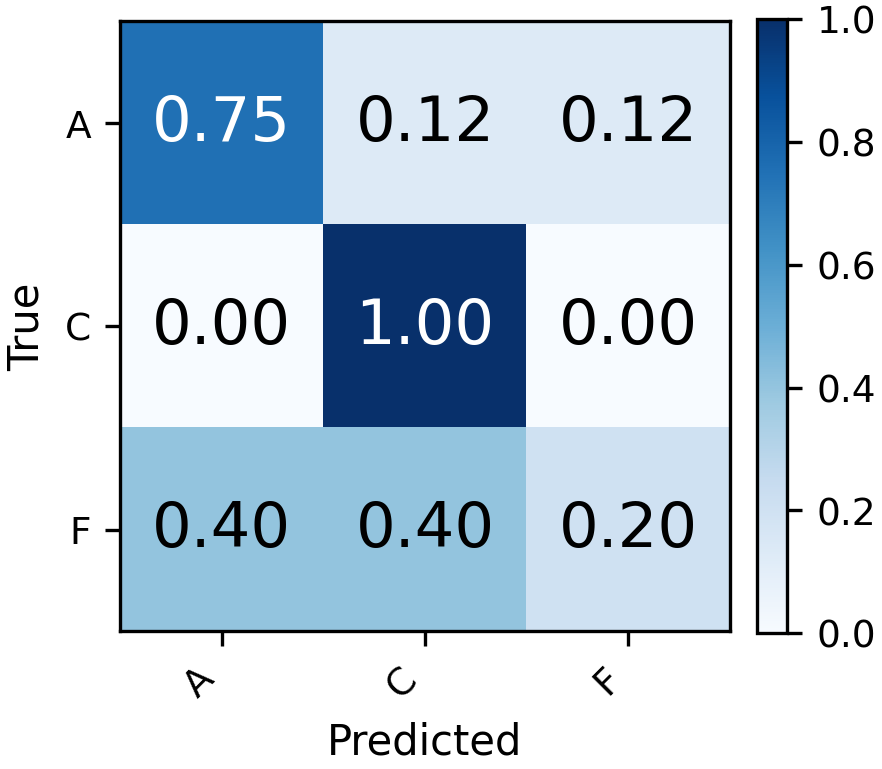}
    \caption{QDA}
    \label{fig:cm_qda}
\end{subfigure}\hfill
\begin{subfigure}[t]{0.48\linewidth}
    \centering
    \includegraphics[scale=0.60]{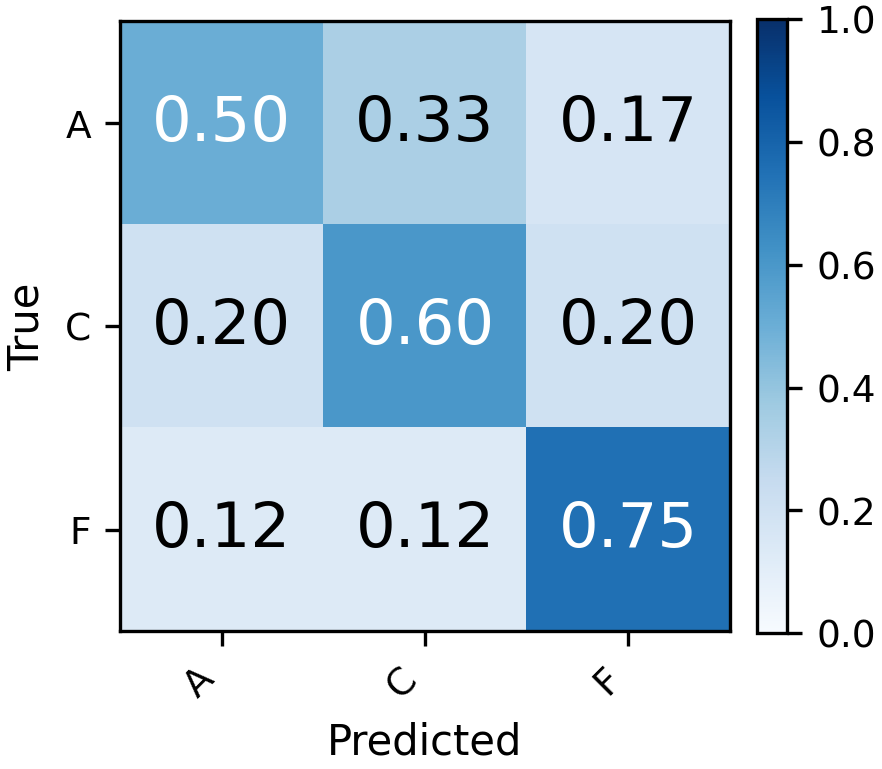}
    \caption{Conformer}
    \label{fig:cm_conformer}
\end{subfigure}

\caption{\textbf{Holdout-set confusion matrices on ADFTD comparing a classical pipeline vs. a Transformer.}
Rows denote ground-truth labels (`A':Alzheimers, `C':Healthy Controls, `F':Dementia) and columns denote predicted labels; darker diagonal entries indicate better class-wise performance. 
\textit{Left:} Quadratic Discriminant Analysis (QDA) trained on aggregated spectral features (Welch-FFT/DWT, channel- and window-averaged) shows stronger separation between Healthy Controls (C) and Alzheimer’s disease (A) on the holdout subjects. 
\textit{Right:} Conformer trained on rs-EEG time-series exhibits reduced true positives for Healthy Controls and Alzheimers. 
Overall, these matrices highlight that feature aggregation can provide a more stable signal-to-noise representation for classical models under limited rs-EEG data.}
\label{fig:QDA_vs_Conformer_CM}
\end{figure}

\subsection{Baseline Comparison and Self-Attention Diagnostics}

In the previous section, we saw that traditional classifiers achieve parity, and sometimes outperform attention-based methods. To emphasize the difference in ability to learn fundamental physiological boundaries between the two approaches, we show the confusion matrix of QDA and Conformer in the ADFTD dataset in Figure \ref{fig:QDA_vs_Conformer_CM}. This dataset comprises of three labels -- Alzheimer's (A), Healthy Control (C), and Frontotemporal Dementia (F). In the spectral domain, healthy brain activity (Control) constitutes a distinct, low-variance state that is perfectly captured by QDA. In contrast, the Conformer fails to recognize this baseline, misclassifying more than one-third of healthy subjects. This indicates that the attention mechanism struggles to extract the consistent feature patterns associated with healthy neural activity in these datasets.
Next, we analyzed the self-attention weights for each test predictions (both correct and incorrect predictions) across all four datasets. We compute the KL divergence between between the attention maps of all pairs of correct (TP,TN) and incorrect (FP, FN) predictions. In Table \ref{tab:KL_Divergence}, we see that a) the values themselves are low, which indicates that the predictive power of the transformer models do not come from the attention mechanism, and b) $KL(Correct||Incorrect) \approx KL(Incorrect||Correct)$, which reveals that the model's internal focus is invariant of diagnostic success.
\begin{table}[!htbp]
\centering
\renewcommand{\arraystretch}{1.15}
\setlength{\tabcolsep}{5pt}
\begin{tabular}{lcc}
\hline
\textbf{Dataset} & \textbf{KL (Correct\textbar\textbar Incorrect)} & \textbf{KL (Incorrect\textbar\textbar Correct)} \\
\hline
ADFTD   & $8.84e-05$ & $8.80e-05$ \\
APAVA   & $4.18e-03$ & $4.16e-03$ \\
TDBrain & $4.87e-03$ & $4.86e-03$ \\
ADHD    & $1.03e-02$ & $1.01e-02$ \\
\hline

\end{tabular}
\caption{\textbf{KL Divergence calculated for all correct vs incorrect and all incorrect vs correct predictions.}{We compute query-key attention weights for all test subjects and compute KL divergence for all pairs of correct and incorrect predictions and report the average scores for KL(Correct || Incorrect) and KL (Incorrect || Correct). \textit{The highest standard deviation found for this experiment was 5.9e-03 for KL(Correct||Incorrect) on the ADHD dataset}.}}
\label{tab:KL_Divergence}
\end{table}

\subsection{Sensitivity analysis of input spectral features}
To investigate whether the parity of attention-based methods are due to inherent stationarity of the oscillatory power and phase-coupling across the dataset, or the denoising effect of the spectral feature engineering, we conducted the following experiment. Time-series from each channel was decomposed into the 5 canonical EEG frequency bands ($\alpha$, $\beta$, $\gamma$, $\delta$, $\theta$). We trained Medformer, which yielded the strongest results across all the attention-based methods on these cleaned, deconvolved signals.

The results, summarized in Table \ref{tab:results_excel} show no appreciable improvement in performance. In fact, as Medformer was provided biologically informed, simpler signals in narrow frequency bands, we see a perfomance plateau. We hypothesize that the self-attention mechanism suffers from ``Attention Dilution'' when presented with multi-band time-series. Instead of focusing on finding discriminative phase-amplitude couplings, the model's capacity is consumed by modeling redundant temporal correlations between highly correlated filtered versions of the same underlying signal. Further, while sub-band decomposition provides explicit physiological priors, it simultaneously quintuples the input dimensionality. In the low-data regime typical of clinical EEG ($N < 100$), this expansion of the hypothesis space likely exacerbates the variance of the Transformer’s attention weights, leading to the observed performance degradation.


\begin{table}[!htbp]
\centering
\renewcommand{\arraystretch}{1.15}
\setlength{\tabcolsep}{8pt} 
\begin{tabular}{lcccccc}
\hline
\textbf{Dataset} & \textbf{Accuracy} & \textbf{Precision} & \textbf{Recall} & \textbf{F1} & \textbf{AUROC} & \textbf{AUPRC} \\
\hline
ADFTD    & $46.88$ & $45.38 $ & $47.53$  & $46.15 $ & $64.06$ & $46.94 $ \\
\hline
APAVA   & $69.88 $ & $70.61 $ & $65.78$ & $65.74$ & $72.39 $ & $72.96 $ \\
\hline
TDBrain & $75.18 $ & $75.78 $ & $75.18$ & $75.04$ & $83.42 $ & $82.75$ \\
\hline
ADHD & $66.23 $ & $66.45 $ & $65.87 $ & $65.43$ & $72.86 $ & $67.64 $ \\

\hline
\end{tabular}

\caption{\textbf{Performance of Medformer with separate signals for $(\delta, \theta, \alpha, \beta, \gamma)  $ bands.} For each of the three rs-EEG datasets, we use the dataset specific configuration provided by Wang et al \cite{c04} and fine-tune augmentations, dropout and weight decay to adjust for validation loss fluctuations. For task-EEG (ADHD) we use the authors' validated baseline configuration. \emph{The highest standard deviation observed across datasets was found to be 3.25 for ADFTD dataset}. This table shows that the self-attention mechanism computes redundant query-key interactions for highly correlated signals across different frequencies. Consequently, the subtle, disease-relevant biomarkers are effectively ``masked'' within the sparse attention space, as the query-key vectors become dominated by these redundant inter-frequency correlations.}
\label{tab:results_excel}
\end{table}

\section{Related Literature}
Both attention and non-attention based methods in time-series classification domain have employed different means to capture the recurring patterns in the signal for classification. In TimesNet \cite{c19}, Wu et al. transform 1D time-series tensors into 2D representations to model intra-period and inter-period variations using convolutional kernels. While their application to medical time-series classification underscores the importance of capturing recurring patterns, the method may be insufficient for specific diagnostic signals. For instance, in EEG signals representing neuro-degenerative disease groups, critical subject-specific periodic variations occur within the 0.5–45 Hz frequency bands \cite{c20,c21,c22}. The standard TimesNet architecture does not explicitly segregate these distinct functional frequency bands, potentially overlooking nuanced pathological markers.
Existing state-of-the-art methods, such as Rocket \cite{c18} and Shapelets \cite{c19}, rely on applying convolutional kernels or extracting discriminative subsequences to capture relevant features for classification. However, in the context of EEG data, standard convolutional kernels often fail to model long-range periodicity effectively. Furthermore, extracting and comparing shapelet subsequences over extended EEG recordings—typically sampled at rates $>100$ Hz—becomes computationally infeasible due to the high dimensionality and length of the time-series.

To model long-range periodicity in medical time-series, state-of-the-art Transformer models employ mechanisms such as Auto-Correlation, frequency-enhanced attention, and cross-channel correlation. However, the Auto-Correlation-based attention used in Autoformer \cite{c05} often fails to capture the specific periodic feature components necessary to discriminate between disease types or healthy controls. Similarly, the frequency-enhanced attention in FEDformer \cite{c07} does not explicitly segregate dominant physiological frequency bands ($\alpha, \beta, \gamma, \delta, \theta$) from overlapping signals; consequently, it remains susceptible to noise in higher frequency ranges (45–100 Hz). More recent architectures, such as Medformer \cite{c08}, incorporate inter-granularity learning and cross-channel correlations. While Yu et al. \cite{c27} address the architectural mismatch between decentralized attention and centralized medical time-series signals, their framework primarily focuses on centralized temporal--channel interactions rather than explicitly modeling disease-specific neuro-pathological spectral biomarkers. Consequently, because the learned representations are not explicitly constrained toward biomarker-relevant spectral components, localized pathological signatures may remain diluted within the broader feature space.


\section{Discussion}

The observed performance parity between traditional classifiers with spectral features, and attention-based architectures operating on time-series data show that the high representational capacity of transformers does not translate to clinical utility for neurophysiological signals. While self-attention is optimized for sparse temporal salience, EEG biomarkers are more often globally distributed evident from table \ref{tab:classic_baselines}. Relative Band powers captured via both time and frequency resolution enabled us to identify spatially stable signals over time.

Furthermore, the inability of transformer-based models to consistently discriminate Healthy Controls with Fronto-temporal Dementia and Alzheimer's Disease in ADFTD reveal that they are unable to capture globally stable spectral features that segregate different disease groups from controls. 
KL divergence results in (Table \ref{tab:KL_Divergence}) suggest that attention weights do not drive classification decisions. This leads to sparse attention distributions that provide limited spectrally informative data for classification. Finding proxy tokens reduces the computational overhead of inter-token interactions in sparse-attention models, but it still does not explicitly uncover the latent biological signatures underlying disease-related signal patterns.
These results suggest that for current clinical data scales, principled parsimony via signal-processing-informed inductive biases remains more robust than domain-agnostic models, advocating for the integration of physics-informed kernels in future neuro-imaging architectures.

%
%

\begin{thebibliography}{8}
\bibitem{c01}
Same, Mohammad Hossein, et al. "Simplified welch algorithm for spectrum monitoring." Applied Sciences 11.1 (2020): 86.

\bibitem{c02}
Radovanović, Miloš, Alexandros Nanopoulos, and Mirjana Ivanović. ``Nearest neighbors in high-dimensional data: The emergence and influence of hubs." Proceedings of the 26th Annual International Conference on Machine Learning. 2009.

\bibitem{c03}
Ghojogh, B., and M. Crowley. ``Linear and quadratic discriminant analysis: Tutorial. arXiv 2019." arXiv preprint arXiv:1906.02590.

\bibitem{c04}
Wang, Yihe, et al. "Medformer: A multi-granularity patching transformer for medical time-series classification." Advances in Neural Information Processing Systems 37 (2024): 36314-36341.

\bibitem{c05}
Wu, Haixu, et al. ``Autoformer: Decomposition transformers with auto-correlation for long-term series forecasting.'' \emph{NeurIPS} 34 (2021): 22419--22430.
\bibitem{c06}
Cleveland, Robert B., et al. "STL: A seasonal-trend decomposition." J. off. Stat 6.1 (1990): 3-73.
\bibitem{c07}
Zhou, Tian, et al. "Fedformer: Frequency enhanced decomposed transformer for long-term series forecasting." International conference on machine learning. PMLR, 2022.

\bibitem{c08}
Kitaev, Nikita, Łukasz Kaiser, and Anselm Levskaya. "Reformer: The efficient transformer." arXiv preprint arXiv:2001.04451 (2020).
\bibitem{c09}
J Escudero et al. Analysis of electroencephalograms in alzheimer’s disease patients with multiscale entropy. Physiological
measurement, 27(11):1091, 2006.
\bibitem{c10}
Hanneke van Dijk, Guido van Wingen, Damiaan Denys, Sebastian Olbrich, Rosalinde van Ruth,
and Martijn Arns. The two decades brainclinics research archive for insights in neurophysiology
(tdbrain) database. Scientific data, 9(1):333, 2022.
\bibitem{c11}
Andreas Miltiadous, Katerina D Tzimourta et al. A dataset of scalp eeg recordings of alzheimer’s disease, frontotemporal dementia and healthy subjects from routine eeg. Data, 8(6):95, 2023.
\bibitem{c12}
Jwo, Dah-Jing, Wei-Yeh Chang, and I-Hua Wu. "Windowing Techniques, the welch method for improvement of Power Spectrum Estimation." Computers, materials \& continua 67.3 (2021).
\bibitem{c13}
Saha S, Baumert M. Intra- and Inter-subject Variability in EEG-Based Sensorimotor Brain Computer Interface: A Review. Front Comput Neurosci. 2020 Jan 21;13:87. doi: 10.3389/fncom.2019.00087. PMID: 32038208; PMCID: PMC6985367.

\bibitem{c14}
Xu, Yilu, Xin Huang, and Quan Lan. "Selective cross-subject transfer learning based on riemannian tangent space for motor imagery brain-computer interface." Frontiers in Neuroscience 15 (2021): 779231.
\bibitem{c15}
Xu, Lichao, et al. "Cross-dataset variability problem in EEG decoding with deep learning." Frontiers in human neuroscience 14 (2020): 103.
\bibitem{c16}
Liang, Shuang, et al. "Adaptive deep feature representation learning for cross-subject EEG decoding." BMC bioinformatics 25.1 (2024): 393.
\bibitem{c17}
Dempster, Angus, François Petitjean, and Geoffrey I. Webb. "ROCKET: exceptionally fast and accurate time series classification using random convolutional kernels." Data Mining and Knowledge Discovery 34.5 (2020): 1454-1495.
\bibitem{c18}
Wu, Haixu, et al. "Timesnet: Temporal 2d-variation modeling for general time series analysis." arXiv preprint arXiv:2210.02186 (2022).
\bibitem{c19}
Cassani, Raymundo, et al. "Systematic review on resting‐state EEG for Alzheimer’s disease diagnosis and progression assessment." Disease markers 2018.1 (2018): 5174815.
\bibitem{c20}
Pal, Anita, et al. "High delta and gamma EEG power in resting state characterise dementia in Parkinson’s patients." Biomarkers in Neuropsychiatry 3 (2020): 100027.
\bibitem{c21}
Durongbhan, Pholpat, et al. "A dementia classification framework using frequency and time-frequency features based on EEG signals." IEEE Transactions on Neural Systems and Rehabilitation Engineering 27.5 (2019): 826-835.
\bibitem{c22}
Y. Song, Q. Zheng, B. Liu and X. Gao, "EEG Conformer: Convolutional Transformer for EEG Decoding and Visualization," in IEEE Transactions on Neural Systems and Rehabilitation Engineering, vol. 31, pp. 710-719, 2023, doi: 10.1109/TNSRE.2022.3230250.
\bibitem{c23}
Zeynali, Mahsa, Hadi Seyedarabi, and Reza Afrouzian. "Classification of EEG signals using Transformer based deep learning and ensemble models." Biomedical Signal Processing and Control 86 (2023): 105130.
\bibitem{c24}
Sarker SR et al. A Hybrid Approach to Attention Deficit Hyperactivity Disorder Detection Leveraging Transformer and XGBoost Models Using XSparseFormerNet. Sci Rep. 2025 Nov 20;15(1):41039. doi: 10.1038/s41598-025-24919-3. PMID: 41266583; PMCID: PMC12635201.
\bibitem{c25}
Swann, Nicole C., et al. "Gamma oscillations in the hyperkinetic state detected with chronic human brain recordings in Parkinson's disease." Journal of Neuroscience 36.24 (2016): 6445-6458.

\bibitem{c26}
Pal, Anita, et al. "High delta and gamma EEG power in resting state characterise dementia in Parkinson’s patients." Biomarkers in Neuropsychiatry 3 (2020): 100027.
\bibitem{c27}
Ali Motie Nasrabadi, Armin Allahverdy, Mehdi Samavati, Mohammad Reza Mohammadi.(2020). "EEG data for ADHD / Control children." Web,

\bibitem{c27}
Yu, Guoqi, et al. "Decentralized Attention Fails Centralized Signals: Rethinking Transformers for Medical Time Series." arXiv preprint arXiv:2602.18473 (2026).
\end{thebibliography}
%

\end{document}